\documentclass{article}
\usepackage{spconf,amsmath,graphicx}

\makeatletter
\def\@maketitle{\newpage
 \null
 \vskip 1.5em
 \begin{center}
  {\large \bf \@title \par}
  \vskip 0.8em
  {\normalsize
   Max A. Nelson$^{1}$, Elif Keles$^{1}$, Eminenur Sen Tasci$^{1}$, Merve Yazol$^{2}$, \\
   Halil Ertugrul Aktas$^{1}$, Ziliang Hong$^{1}$, Andrea Mia Bejar$^{1}$, \\
   Gorkem Durak$^{1}$, Oznur Leman Boyunaga$^{2}$, Ulas Bagci$^{1}$
   \par}
  \vskip 0.5em
  {\small
   $^{1}$Northwestern University, Dept. of Radiology, Chicago, IL, USA \\
   $^{2}$Gazi University, Dept. of Radiology, Ankara, T\"{u}rkiye \\
   \vspace{1mm}
   Correspondence: maxnelson2025 [at] u [dot] northwestern [dot] edu
  }
 \end{center}
 \par
 \vskip 1em}
\makeatother

\title{Upstream Probabilistic Meta-Imputation for Multimodal Pediatric Pancreatitis Classification}
\begin{document}
\maketitle
\begin{abstract}
Pediatric pancreatitis is a progressive and debilitating inflammatory condition, including acute pancreatitis and chronic pancreatitis, that presents significant clinical diagnostic challenges. Machine learning-based methods also face diagnostic challenges due to limited sample availability and multimodal imaging complexity. To address these challenges, this paper introduces Upstream Probabilistic Meta‑Imputation (UPMI), a light‑weight augmentation strategy that operates upstream of a meta‑learner in a low‑dimensional meta‑feature space rather than in image space. Modality-specific logistic regressions (T1W and T2W MRI radiomics) produce probability outputs that are transformed into a 7‑dimensional meta‑feature vector. Class‑conditional Gaussian mixture models (GMMs) are then fit within each cross‑validation fold to sample synthetic meta‑features that, combined with real meta‑features, train a Random Forest (RF) meta‑classifier. On 67 pediatric subjects with paired T1W/T2W MRIs, UPMI achieves a mean AUC of 0.908 $\pm$ 0.072, a $\sim$5\% relative gain over a real‑only baseline (AUC 0.864 $\pm$ 0.061). 
\end{abstract}
\begin{keywords}
Pediatric pancreatitis, meta-learner, synthetic data augmentation, modality fusion, small sample learning
\end{keywords}

\section{Introduction}
\label{sec:intro}
\begin{figure}[h!]
\centering
\includegraphics[width=0.8\linewidth]{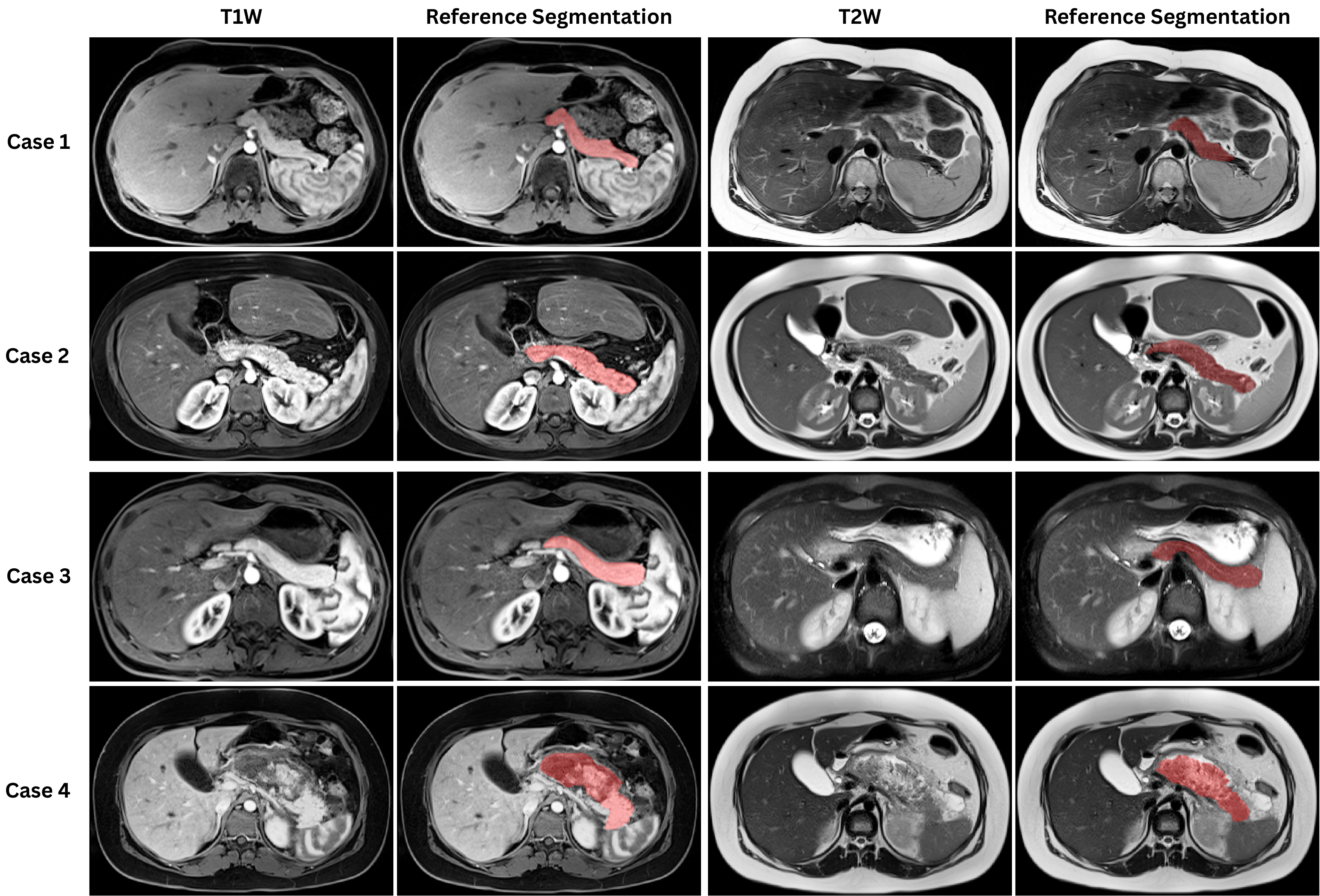}
\caption{Representative cases from the pediatric pancreas MRI dataset showing T1W and T2W sequences. Reference pancreas segmentations are highlighted in red. First two rows (Cases 1-2): healthy control subjects; last two rows (Cases 3-4): pancreatitis patients from our cohort.}
\label{fig:segmentation_viz}
\end{figure}
Pediatric pancreatitis includes acute (AP) and chronic (CP) forms and has increased to \textbf{roughly 13 cases per 100,000 children} annually~\cite{uc2019pancreatitis,tian2022etiology}. AP involves sudden pancreatic inflammation that can improve with treatment; however, some children develop CP, characterized by permanent structural changes and loss of function \cite{uc2019pancreatitis,morinville2012definitions}. MRI is central to assessment because it provides radiation‑free, high‑contrast, complementary information from T1‑weighted (T1W) and T2‑weighted (T2W) sequences. Yet the pediatric pancreas is small, changes shape with age, and often has low contrast with adjacent tissue; motion and peripancreatic fluid collections further complicate interpretation~\cite{keles2025pediatric}. These factors, combined with limited cohort sizes, motivate methods that can reliably fuse multimodal MRI under small‑sample constraints~\cite{keles2025pediatric,kemp2025practical}.

\begin{figure*}[t]
    \centering
    \includegraphics[width=\linewidth]{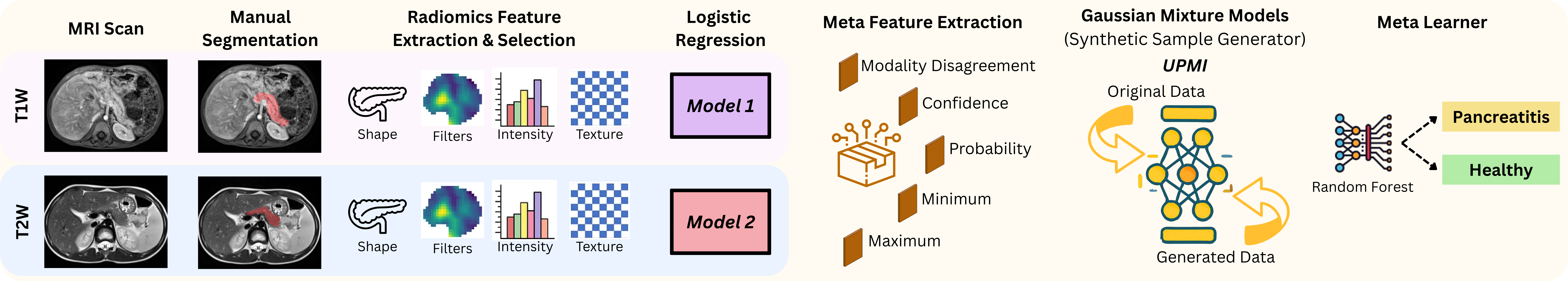}
    \caption{Multi-modality pipeline for pediatric pancreatitis classification with the proposed method UPMI.}
\label{fig:Model}
\end{figure*}

\textbf{Multimodal fusion} commonly uses feature concatenation or stacking, where a meta‑learner combines modality‑specific predictions. These approaches can help, but overfit in data‑constrained settings~\cite{stahlschmidt2022multimodal,sadasivuni2022fusion}, typical of pediatric MRI. Image‑space augmentation with GANs/VAEs/diffusion can expand data, but it is computationally heavy and may generate unrealistic images that do not transfer to clinical signals. UPMI addresses both issues by augmenting in meta‑feature space—the decision layer—where distributions are easier to model, computation is modest, and leakage can be controlled via fold‑localized sampling~\cite{kebaili2023deep,skandarani2023gans,sizikova2024synthetic}.

\textbf{Our Contributions.} To address these challenges, we introduce Upstream Probabilistic Meta-Imputation (UPMI), a computationally light and interpretable strategy that augments a low-dimensional meta‑feature space derived from cross‑validated outputs of modality‑specific base classifiers to improve learning under small‑sample, multimodal conditions. The key technical contribution is a fold‑localized, class‑conditional Gaussian mixture sampler that generates validity‑constrained synthetic meta‑features within each training fold and combines them with real meta‑features to train a Random‑Forest meta‑classifier while minimizing information leakage. By operating at the decision layer rather than the pixel level, the approach provides principled, leakage‑aware augmentation with minimal computational overhead. Clinically, the study delivers a preliminary dual‑sequence (T1W/T2W) MRI classifier for pediatric pancreatitis that attains promising discrimination (mean AUC $\approx$ 0.91) on a small single‑center cohort, highlighting the feasibility of lightweight meta‑space augmentation for rare pediatric disease imaging.
\section{Related Works}
Recent advances in radiomics-based machine learning techniques have heightened interest in their application for detecting pancreatic diseases, yielding significant improvements in early diagnosis. Studies have investigated early detection of pancreatic cancer using various machine learning classifiers with radiomic features extracted from pre-diagnostic abdominal CT scans \cite{mukherjee2022radiomics}. These statistical models have demonstrated success in clinical decision-making, particularly in risk stratification.

Yao et al. developed a hybrid framework that combines deep learning and radiomics for the risk stratification of IPMN using multimodal MRI. They implemented automatic pancreas segmentation and a two-channel deep neural network for feature extraction, achieving an accuracy of 0.82 in a multicenter study involving 246 MRI scans, which markedly improved detection compared to traditional guidelines \cite{yao2023radiomics}.

Jiang et al. introduced a multi-learner meta-learning (MAML) framework that merges transfer learning and metric learning, enhanced with real-time and soft labels, reaching state-of-the-art performance across several few-shot benchmarks \cite{jiang2022multi}.
López-Ríos et al. demonstrated that incorporating conditional GAN-generated MRI into the training process can consistently improve classification results for multiple sclerosis without compromising the biological significance of feature patterns \cite{lopez2025assessing}. These works indicate that both meta-learning and the generative approach can effectively boost performance in conditions of limited data.

\section{Methods}
\label{sec:method}

\subsection{Data Aggregation and Annotation}
\label{ssec:data}

 This retrospective study involved 67 children (aged 2-19 years) with corresponding T1W and T2W abdominal MRIs (134 images total) obtained using a combination of 1.5T and 3T Siemens Aera/Verio machines. Study subjects in the healthy control group (class 0, 43 patients; 64.2\%) present with uncomplicated pancreases, instead of undergoing imaging for conditions such as hypo-echoic liver lesions. The remaining 24 patients (class 1, 35.8\%) have been previously diagnosed with either AP or  CP based on clinical criteria. Because of sample size limitations, we merged both acute and chronic pancreatitis into a single positive class. 

Expert radiologists manually annotated segmentation masks for each modality \cite{keles2025pediatric}. Mask veracity is further validated through quantitative comparison against automated segmentations generated by the state-of-the-art PanSegNet model, which has extensively documented leading performance on adult pancreas segmentation benchmarks \cite{keles2025pediatric,zhang2025large,sonoda2025comparison}. The resulting masks, partially visualized in Fig.~\ref{fig:segmentation_viz}, precisely delineate the pancreas from surrounding structures and define the region of interest for radiomics-based feature extraction.

\subsection{Radiomics Feature Extraction and Selection}
\label{ssec:radiomics}

For each modality (T1W and T2W) independently, we extract all available features through the PyRadiomics package with wavelet and Laplacian of Gaussian enabled. To mitigate dimensionality concerns, we adhere to an aggressive, four-stage feature selection process. We first discard features with variance below 0.01, then apply a correlation filter in which, for any pair with Pearson correlation $|\rho| > 0.95$, we keep only the feature with higher mutual information with respect to the class label. From the remaining features, we select the top 20 by mutual information $I(X;Y)$. Lastly, we train an RF on these candidate features, retaining only the 10 most important by mean decrease impurity. All feature selection steps are performed within the training folds of cross-validation to prevent data leakage.

\subsection{Baseline Logistic Regression Classifiers}
\label{ssec:baseline_lr}

We train independent L2-regularized logistic regression classifiers for T1W and T2W modalities separately. Each model optimizes the regularized binary cross-entropy loss:

\begin{equation}
\mathcal{L}(\mathbf{w}, b)
= \frac{1}{2C}\|\mathbf{w}\|_2^2
  + \sum_{i=1}^{n} w_i \,\ell(y_i, \hat{y}_i),
\label{eq:lr_loss}
\end{equation}
\noindent where $\mathbf{w}$ and $b$ denote the feature weights and bias, $C = 1.0$ is the
inverse regularization strength, $\hat{y}_i = \sigma(\mathbf{w}^\top
\mathbf{x}_i + b)$ is the predicted probability under the logistic sigmoid
$\sigma(\cdot)$, and $\ell(y_i, \hat{y}_i) = -y_i \log \hat{y}_i -
(1-y_i)\log(1-\hat{y}_i)$ is the per-sample cross-entropy loss. The
sample-specific weights $w_i$ address class imbalance (43 control, 24 AP/CP;
ratio 1.8:1).


\subsection{Synthetic Meta-Feature Extraction}
\label{ssec:metafeatures}

For each patient $i$ in the test set of outer fold $k$, let $p_i^{T1}$ and $p_i^{T2}$ denote the predicted probabilities for class 1 from the T1W and T2W classifiers. We then define a 7-dimensional meta-feature vector:

\begin{equation}
\begin{aligned}
\mathbf{m}_i = [\, &p_i^{T1},\, p_i^{T2},\, c_i^{T1},\, c_i^{T2},\, d_i,\\
&\max(p_i^{T1}, p_i^{T2}),\, \min(p_i^{T1}, p_i^{T2}) \,],
\end{aligned}
\end{equation}
\noindent where $c_i^{T1} = \max(p_i^{T1}, 1-p_i^{T1})$ is T1W confidence, $c_i^{T2} = \max(p_i^{T2}, 1-p_i^{T2})$ is T2W confidence, and $d_i = |p_i^{T1} - p_i^{T2}|$ is modality disagreement. This meta-feature vector attempts to provide a compact representation of individual modality predictions while capturing any inter-modality relationships. Each training patient in outer fold $k$ receives predictions from base models trained on a subset of the training data that excludes that patient. This ensures all meta-features in both training and test are derived from out-of-fold predictions, maintaining strict separation at every level of the pipeline.

\subsection{Synthetic Meta-Feature Generation}
\label{ssec:synthetic}

Given our limited availability of real meta-features (approximately 53 training samples per fold), we seek to augment the training set by sampling features using Gaussian Mixture Models (GMM) upstream of the baseline classifier (visualized in Fig.~\ref{fig:Model}). For each class $c \in \{0, 1\}$, we fit a GMM with $K=2$ components to the meta-features of that class:

\begin{equation}
p(\mathbf{m} | c) = \sum_{k=1}^{K} \pi_k^c \mathcal{N}(\mathbf{m} | \boldsymbol{\mu}_k^c, \boldsymbol{\Sigma}_k^c),
\end{equation}
where $\pi_k^c$ are mixture weights, $\boldsymbol{\mu}_k^c$ are component means, and $\boldsymbol{\Sigma}_k^c$ are full covariance type matrices. To generate synthetic meta-features, we draw samples from our fitted GMM: $\tilde{\mathbf{m}} \sim p(\mathbf{m} | c)$. We then apply
feature-type validity constraints derived from the reference meta-feature definition in Section~\ref{ssec:metafeatures}: probability-like dimensions are clipped to $[0,1]$ and derived entries (confidence, disagreement, max and min probability) are computed accordingly. 

\subsection{Meta-Learner Training with Augmented Data}
\label{ssec:metalearner}

As the final layer in our pipeline, we train a RF meta-learner using both the original and imputed (synthetic) meta features. The RF is configured with 100 trees and maximum depth 5 to prevent overfitting. Features are standardized with zero mean and unit variance before training.
The RF ensemble prediction aggregates $T=100$ decision trees via
majority voting:
\begin{equation}
\hat{y}(\mathbf{m}) = \operatorname{mode}\{h_t(\mathbf{m})\}_{t=1}^{T},
\end{equation}
and the class probability is estimated as
\begin{equation}
p(y = c \mid \mathbf{m}) = \frac{1}{T} \sum_{t=1}^{T} \mathbf{1}\big[h_t(\mathbf{m}) = c\big],
\end{equation}
where each tree $h_t$ is trained on a bootstrap sample from the augmented meta-feature
dataset $\mathcal{D}_{\text{aug}} = \mathcal{D}_{\text{real}} \cup
\mathcal{D}_{\text{synth}}$ with maximum depth $d_{\max} = 5$ for
regularization.

To determine the optimal amount of synthetic data, we conduct an ablation study by varying the percent of synthetic data, measured relative to the real data quantity, included in the augmented dataset. We test five scenarios, detailed in Table~\ref{tab:ablation}, and balance synthetic samples across classes.
\section{Results}
\label{sec:results}

\subsection{Ablation Study Results}
\label{ssec:ablation}

Table \ref{tab:ablation} summarizes the meta-learner's classification performance across synthetic data scenarios. We observe a generally positive dose-response relationship, with superior performance at 200\% synthetic data. This corresponds to an AUC of $0.908 \pm 0.072$, with $+0.044$ absolute AUC increase and and $+5.1\%$ relative improvement. The relevant mean ROC curve is visualized in Fig.~\ref{fig:rocconf}a.

\begin{table}[h]
\centering
\caption{RF Meta-Learner Performance Across Synthetic Data Scenarios (Mean $\pm$ Std over 5 folds)}
\label{tab:ablation}
\begin{tabular}{l|c|c|c}
\hline
\textbf{Scenario} & \textbf{N Synth} & \textbf{AUC} & \textbf{F1} \\
\hline
Real only & 0 & 0.864 $\pm$ 0.061 & 0.775 $\pm$ 0.060 \\
Real+25\% & 13 & 0.862 $\pm$ 0.055 & 0.769 $\pm$ 0.055 \\
Real+50\% & 26 & 0.881 $\pm$ 0.071 & 0.788 $\pm$ 0.062 \\
Real+100\% & 53 & 0.901 $\pm$ 0.077 & 0.805 $\pm$ 0.067 \\
\textbf{Real+200\%} & \textbf{106} & \textbf{0.908 $\pm$ 0.072} & \textbf{0.810 $\pm$ 0.067} \\
\hline
\end{tabular}
\end{table}

\subsection{Statistical Analysis}
\label{ssec:statistics}

A paired t-test comparing the optimal scenario (Real + 200\% synthetic) to the
real-only baseline yields p = 0.051, suggesting a strong trend toward
improvement but narrowly missing the conventional 0.05 threshold. The mean AUC
gain is +0.043, indicative of a large effect size (Cohen's $d = 1.24$,
i.e., $> 0.8$). A nonparametric bootstrap of the fold-wise AUC differences
produces a 95\% confidence interval of [0.0167, 0.0700], which notably excludes zero,
and 4 of 5 folds show higher AUC under the augmented setting. Taken together, these results indicate a substantial and consistently positive effect, with the
borderline t-test p-value plausibly explained by the limited number of paired observations.

\subsection{Synthetic Data Quality Validation}
\label{ssec:validation}

\begin{figure}[h]
\centering
\includegraphics[width=1\linewidth]{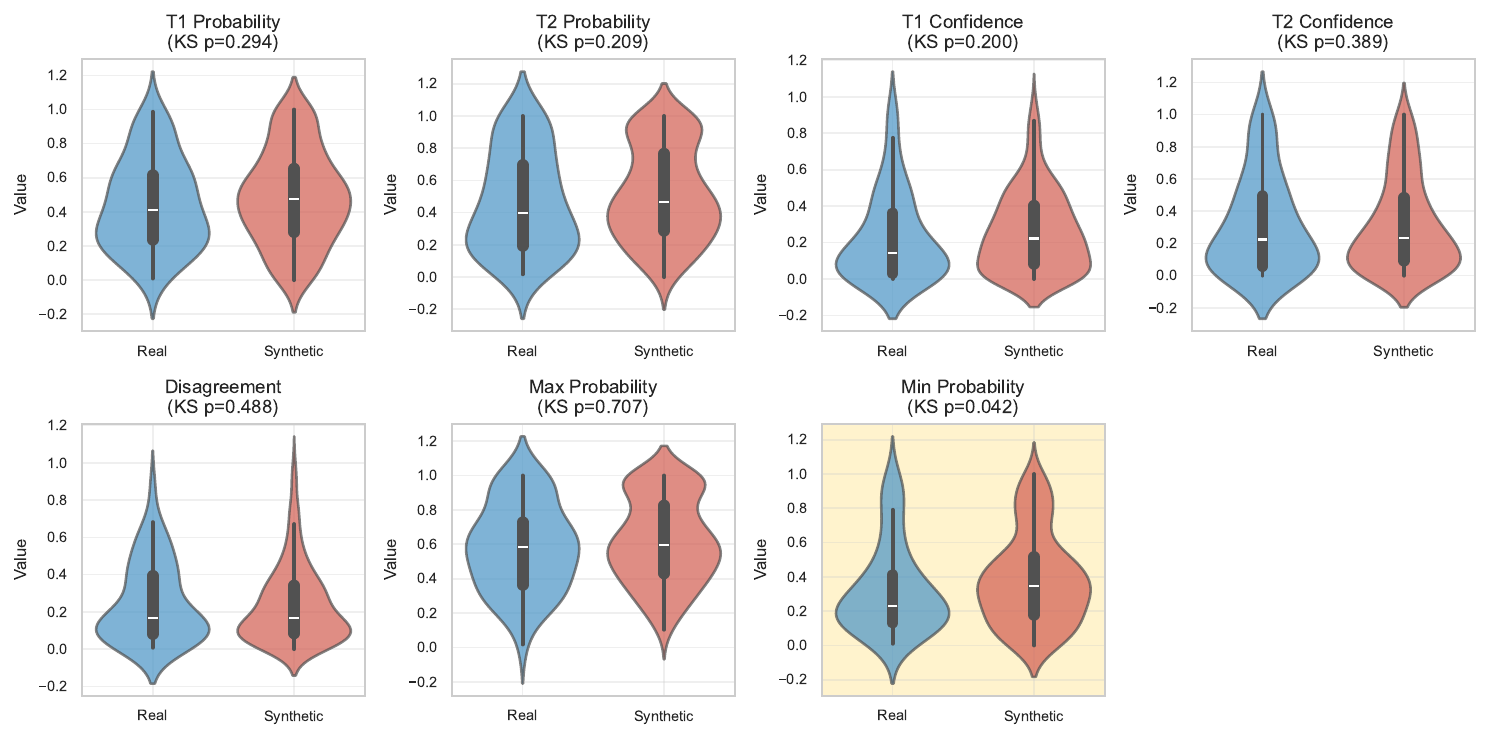}
\caption{Violin plot comparison of real and GMM-sampled meta-feature distributions. Titles report two-sample KS p-values; the borderline significant shift feature is shaded.}
\label{fig:metafeature_violin}
\end{figure}

Two-sample Kolmogorov-Smirnov tests comparing real and synthetic meta-feature distributions reveal 6/7 features have statistically similar distributions (p $>$ 0.05, depicted in Fig.~\ref{fig:metafeature_violin}), with an average p-value 0.333, indicating satisfactory synthetic data quality. Only \texttt{min\_probability} shows a modest but statistically significant divergence (p = 0.042). However, as this isolated shift among seven features is small in magnitude, we regard the overall agreement between real and synthetic meta-features as acceptable.

\begin{figure}[t]
\centering
\includegraphics[width=1\linewidth]{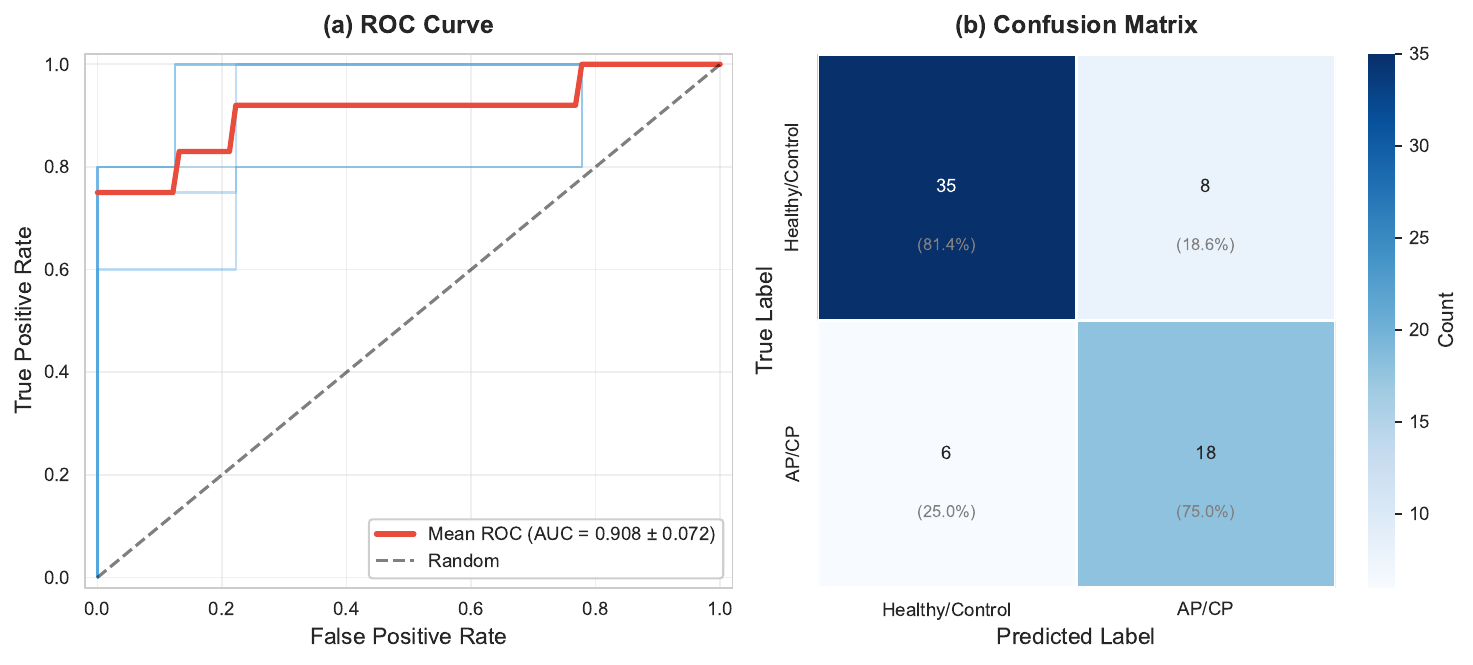}
\caption{(a) Mean ROC curve across 5 folds, (b) Aggregated confusion matrix; both correspond to optimal config.}
\label{fig:rocconf}
\end{figure}

\subsection{Error Analysis}
At the default decision threshold, the model correctly identifies 18/24 AP/CP cases (75.0\% sensitivity) while correctly identifying 35/43 healthy cases (81.4\% specificity; Fig.~\ref{fig:rocconf}b). Our error analysis revealed that both false negatives and false positives exhibited greater disagreement between the model and the reader compared to correctly classified cases, highlighting the challenges associated with borderline imaging presentations. False negatives were often associated with subtle or atypical disease manifestations, making them difficult for the model to detect. Meanwhile, false positives arose from benign features, like peripancreatic fat stranding, that the model misinterpreted as disease. These findings highlight the limitations of multimodal machine learning systems in recognizing subtle signs of disease and underscore the need for improved contextual understanding and more diverse training data.

\begin{figure}[h]
\centering
\includegraphics[width=1\linewidth]{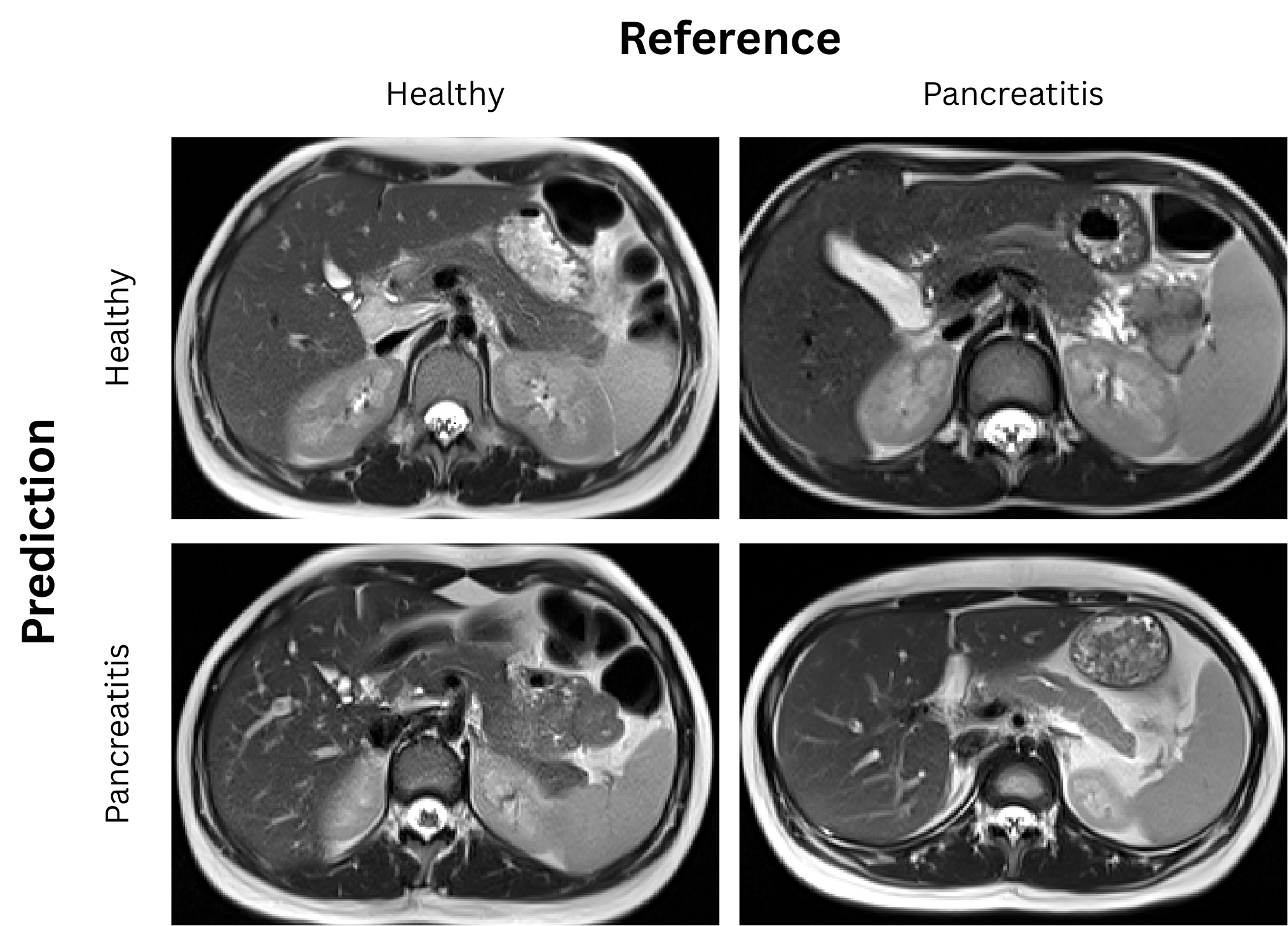}
\caption{Representative misclassifications arranged as a confusion‑matrix grid (rows=reference; columns=prediction) using axial T1W/T2W slices. False negatives typically reflect subtle or atypical pancreatitis, whereas false positives often arise from peripancreatic fat stranding; true cases are shown for context. See aggregate counts in Figure 4b.}
\label{fig:scanconf}
\end{figure}

\section{Conclusion}
\label{sec:discussion}
We presented Upstream Probabilistic Meta Imputation (UPMI), a lightweight meta-learning fortification technique for multimodal medical image classification with limited samples. By operating in meta-feature space and using GMM-based augmentation, UPMI achieves $\sim$+5.0\% relative AUC improvement with large effect size (Cohen's d = 1.24), demonstrating practical utility for small-sample scenarios. The approach is computationally efficient, interpretable, and generalizable to other multimodal prediction tasks. Future work includes validation on larger datasets, an ablation study on meta-learner configuration, and further exploration of alternative sampling techniques such as logistic-normal and Beta mixtures. 

\section{Compliance with Ethical Standards}
\label{sec:ethics}
This study was performed in line with the principles of the Declaration of Helsinki. All patient data were de-identified prior to analysis. Approval was granted by the Gazi University Institutional Review Board.

\section{Acknowledgments}
\label{sec:acknowledgments}
This work was partially supported by NIH R01-CA246704.


\begin{thebibliography}{10}

\bibitem{uc2019pancreatitis}
Aliye Uc and Sohail~Z Husain,
\newblock ``Pancreatitis in children,''
\newblock {\em Gastroenterology}, vol. 156, no. 7, pp. 1969--1978, 2019.

\bibitem{tian2022etiology}
Guo Tian, Lu~Zhu, Shuochun Chen, Qiyu Zhao, and Tian'an Jiang,
\newblock ``Etiology, case fatality, recurrence, and severity in pediatric acute pancreatitis: a meta-analysis of 48 studies,''
\newblock {\em Pediatric Research}, vol. 91, no. 1, pp. 56--63, 2022.

\bibitem{morinville2012definitions}
Veronique~D Morinville, Sohail~Z Husain, Harrison Bai, Bradley Barth, Rabea Alhosh, Peter~R Durie, Steven~D Freedman, Ryan Himes, Mark~E Lowe, John Pohl, et~al.,
\newblock ``Definitions of pediatric pancreatitis and survey of present clinical practices,''
\newblock {\em Journal of pediatric gastroenterology and nutrition}, vol. 55, no. 3, pp. 261--265, 2012.

\bibitem{keles2025pediatric}
Elif Keles, Merve Yazol, Gorkem Durak, Ziliang Hong, Halil~Ertugrul Aktas, Zheyuan Zhang, Linkai Peng, Onkar Susladkar, Necati Guzelyel, Oznur~Leman Boyunaga, et~al.,
\newblock ``Pediatric pancreas segmentation from mri scans with deep learning,''
\newblock {\em Pancreatology}, 2025.

\bibitem{kemp2025practical}
Justine~M Kemp, Adarsh Ghosh, Jonathan~R Dillman, Rekha Krishnasarma, Mary~Kate Manhard, Aaryani Tipirneni-Sajja, Utsav Shrestha, Andrew~T Trout, and Cara~E Morin,
\newblock ``Practical approach to quantitative liver and pancreas mri in children,''
\newblock {\em Pediatric Radiology}, vol. 55, no. 1, pp. 36--57, 2025.

\bibitem{stahlschmidt2022multimodal}
S{\"o}ren~Richard Stahlschmidt, Benjamin Ulfenborg, and Jane Synnergren,
\newblock ``Multimodal deep learning for biomedical data fusion: a review,''
\newblock {\em Briefings in bioinformatics}, vol. 23, no. 2, pp. bbab569, 2022.

\bibitem{sadasivuni2022fusion}
Sudarsan Sadasivuni, Monjoy Saha, Neal Bhatia, Imon Banerjee, and Arindam Sanyal,
\newblock ``Fusion of fully integrated analog machine learning classifier with electronic medical records for real-time prediction of sepsis onset,''
\newblock {\em Scientific reports}, vol. 12, no. 1, pp. 5711, 2022.

\bibitem{kebaili2023deep}
Aghiles Kebaili, J{\'e}r{\^o}me Lapuyade-Lahorgue, and Su~Ruan,
\newblock ``Deep learning approaches for data augmentation in medical imaging: a review,''
\newblock {\em Journal of imaging}, vol. 9, no. 4, pp. 81, 2023.

\bibitem{skandarani2023gans}
Youssef Skandarani, Pierre-Marc Jodoin, and Alain Lalande,
\newblock ``Gans for medical image synthesis: An empirical study,''
\newblock {\em Journal of Imaging}, vol. 9, no. 3, pp. 69, 2023.

\bibitem{sizikova2024synthetic}
Elena Sizikova, Andreu Badal, Jana~G Delfino, Miguel Lago, Brandon Nelson, Niloufar Saharkhiz, Berkman Sahiner, Ghada Zamzmi, and Aldo Badano,
\newblock ``Synthetic data in radiological imaging: current state and future outlook,''
\newblock {\em BJR| Artificial Intelligence}, vol. 1, no. 1, pp. ubae007, 2024.

\bibitem{mukherjee2022radiomics}
Sovanlal Mukherjee, Anurima Patra, Hala Khasawneh, Panagiotis Korfiatis, Naveen Rajamohan, Garima Suman, Shounak Majumder, Ananya Panda, Matthew~P Johnson, Nicholas~B Larson, et~al.,
\newblock ``Radiomics-based machine-learning models can detect pancreatic cancer on prediagnostic computed tomography scans at a substantial lead time before clinical diagnosis,''
\newblock {\em Gastroenterology}, vol. 163, no. 5, pp. 1435--1446, 2022.

\bibitem{yao2023radiomics}
Lanhong Yao, Zheyuan Zhang, Ugur Demir, Elif Keles, Camila Vendrami, Emil Agarunov, Candice Bolan, Ivo Schoots, Marc Bruno, Rajesh Keswani, et~al.,
\newblock ``Radiomics boosts deep learning model for ipmn classification,''
\newblock in {\em International Workshop on Machine Learning in Medical Imaging}. Springer, 2023, pp. 134--143.

\bibitem{jiang2022multi}
Hongyang Jiang, Mengdi Gao, Heng Li, Richu Jin, Hanpei Miao, and Jiang Liu,
\newblock ``Multi-learner based deep meta-learning for few-shot medical image classification,''
\newblock {\em IEEE Journal of Biomedical and Health Informatics}, vol. 27, no. 1, pp. 17--28, 2022.

\bibitem{lopez2025assessing}
Erick~Eduardo L{\'o}pez-R{\'\i}os and Francisco~J Alvarez-Padilla,
\newblock ``Assessing ai-augmented training for multiple sclerosis classification in a basal ganglia radiomics model,''
\newblock {\em BMC Medical Imaging}, vol. 25, no. 1, pp. 1--13, 2025.

\bibitem{zhang2025large}
Zheyuan Zhang, Elif Keles, Gorkem Durak, Yavuz Taktak, Onkar Susladkar, Vandan Gorade, Debesh Jha, Asli~C Ormeci, Alpay Medetalibeyoglu, Lanhong Yao, et~al.,
\newblock ``Large-scale multi-center ct and mri segmentation of pancreas with deep learning,''
\newblock {\em Medical image analysis}, vol. 99, pp. 103382, 2025.

\bibitem{sonoda2025comparison}
Yuki Sonoda, Shota Fujisawa, Mariko Kurokawa, Wataru Gonoi, Shouhei Hanaoka, Takeharu Yoshikawa, and Osamu Abe,
\newblock ``Comparison of publicly available artificial intelligence models for pancreatic segmentation on t1-weighted dixon images,''
\newblock {\em Japanese Journal of Radiology}, pp. 1--7, 2025.

\end{thebibliography}
\end{document}